\documentclass[11pt]{article}
\usepackage[final]{acl}
\usepackage{times}
\usepackage{latexsym}
\usepackage[T1]{fontenc}
\usepackage[utf8]{inputenc}
\usepackage{microtype}
\usepackage{inconsolata}
\usepackage{graphicx}
\usepackage{multicol}
\usepackage{multirow}
\usepackage{booktabs}
\usepackage{makecell}
\usepackage{colortbl}
\usepackage[cmyk]{xcolor}
\usepackage{amssymb}
\usepackage{amsmath}
\usepackage{diagbox}
\usepackage[ruled,linesnumbered]{algorithm2e} 
\usepackage{hyperref}
\usepackage{pifont}

\title{HiEdit: Lifelong Model Editing with Hierarchical \\ Reinforcement Learning}

\author{Yangfan Wang\textsuperscript{1}, Tianyang Sun\textsuperscript{1}, Chen Tang\textsuperscript{4}, Jie Liu\textsuperscript{1,3}, Wei Cai\textsuperscript{2,3}\footnotemark[1], Jingchi Jiang\textsuperscript{1,3}\footnotemark[1]  \\
  \textsuperscript{1}Harbin Institute of Technology, \textsuperscript{2}Beidahuang Information Co., Ltd. \\
  \textsuperscript{3}State Key Laboratory of Smart Farm Technologies and Systems \\ \textsuperscript{4}AI Research Center, Midea Group (Shanghai) Co., Ltd. \\
  \texttt{\{yf.wang,25s103212\}@stu.hit.edu.cn}, \texttt{travistang@foxmail.com} \\
  \texttt{icaiweig@gmail.com}, \texttt{\{jieliu,jiangjingchi\}@hit.edu.cn}}

\begin{document}
\maketitle

\renewcommand{\thefootnote}{\fnsymbol{footnote}} 
\footnotetext[1]{Corresponding author.} 
\renewcommand*{\thefootnote}{\arabic{footnote}}

\begin{abstract}

Lifelong model editing (LME) aims to sequentially rectify outdated or inaccurate knowledge in deployed LLMs while minimizing side effects on unrelated inputs. However, existing approaches typically apply parameter perturbations to a static and dense set of LLM layers for all editing instances. This practice is counter-intuitive, as we hypothesize that different pieces of knowledge are stored in distinct layers of the model. Neglecting this layer-wise specificity can impede adaptability in integrating new knowledge and result in catastrophic forgetting for both general and previously edited knowledge. To address this, we propose \textbf{HiEdit}, a hierarchical reinforcement learning framework that adaptively identifies the most knowledge-relevant layers for each editing instance. By enabling dynamic, instance-aware layer selection and incorporating an intrinsic reward for sparsity, HiEdit achieves precise, localized updates. Experiments on various LLMs show that HiEdit boosts the performance of the competitive RLEdit by an average of 8.48\% with perturbing only half of the layers per edit. Our code is available at: \href{https://github.com/yangfanww/hiedit}{https://github.com/yangfanww/hiedit}.

\end{abstract}

\section{Introduction}

\begin{figure}[ht]
  \centering
  \includegraphics[width=\linewidth]{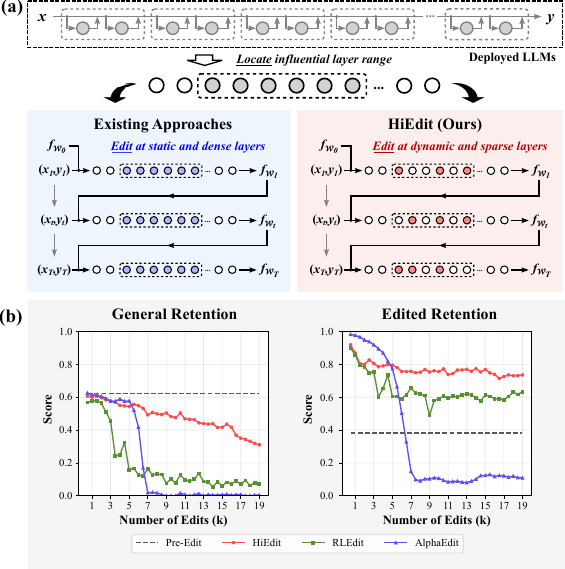}
  \caption {(a) provides a comparison between HiEdit and existing LME approaches. (b) illustrates the ability of various methods to retain general and previously edited knowledge during long-range sequential editing.}
  \label{fig:open}
\end{figure}

Large language models (LLMs) acquire vast knowledge through extensive training on pre-training corpora \citep{petroni2019language, brown2020language}. However, the knowledge stored in deployed LLMs inevitably becomes outdated or inaccurate over time \citep{de-cao-etal-2021-editing, lazaridou2021mind}. Given the actual need for continuous knowledge updates, retraining LLMs from scratch is both costly and impractical. Lifelong model editing (LME) has emerged as a promising solution, aiming to continuously perform targeted updates on deployed LLMs without degrading performance on general or previously edited knowledge \citep{hartvigsen2023aging}. Recent advances in LME typically follow a ``locating-then-editing'' paradigm \citep{meng2022locating}, which involves first identifying influential parameters $\mathcal{W}$, and then editing them by introducing perturbations $\tilde{\nabla}_{\mathcal{W}}$. Current methods can be categorized into two primary approaches: one \citep{meng2022locating,meng2022mass,fang2024alphaedit} calculates closed-form solutions through repetitive matrix operations for parameter updates, which can be costly and error-prone; the other \citep{mitchell2021fast,tan2023massive,li2025reinforced} employs hypernetworks to generate parameter updates efficiently, leveraging low-rank decomposition theory of gradients and parameters. Notably, RLEdit \citep{li2025reinforced} pioneers the framing of LME as a reinforcement learning (RL) task, treating parameter updates as actions to capture the long-range dependencies of editing trajectories, enabling lifelong model editing across tens of thousands of knowledge instances.

However, a critical limitation persists in existing approaches: \textit{they perturb parameters at static and dense LLM layers for all knowledge instances}. Recent research indicates that distinct knowledge activates different components\footnote{For ease of understanding, we use the concepts of ``component'', ``layer module'', and ``layer'' interchangeably.} within LLMs, with only a small subset associated with specific knowledge \citep{yao2024knowledge, wang2024roselora}. Consequently, applying perturbations to static and dense layers indiscriminately leads to the over-modification of parameters, impeding the precise integration of new knowledge and causing the forgetting of existing information. Theoretically, enforcing updates on irrelevant layers unnecessarily constrains the optimization landscape of hypernetworks, resulting in suboptimal solutions. Experimentally, this lack of adaptive layer selection manifests as severe catastrophic forgetting. As illustrated in Figure~\ref{fig:open}(b), when performing 20,000 sequential edits from the ZsRE dataset on Llama-3-8B model, the ability of the LLM to retain both general and previously edited knowledge begins to decline significantly after just 5,000 sequential edits using existing LME methods.
  
To address these limitations, we propose HiEdit, a novel solution using hierarchical reinforcement learning (HRL) to facilitate adaptive and localized LME. We frame LME as a hierarchical decision-making process, explicitly decoupling the complex editing task into two subtasks: layer selection (where to edit) and parameter updating (how to edit). This hierarchical structure transforms the flat and intractable action space into a manageable structured framework, allowing the editor to dynamically pinpoint appropriate components for different knowledge. Specifically, the high-level hypernetwork operates as a manager, leveraging gradient signals to adaptively output a layer importance distribution. The low-level hypernetwork then acts as a worker, generating effective parameter updates for the selected components. To enable end-to-end training across discrete decisions, we bridge the high-level and low-level policies to ensure efficient gradient flow. Furthermore, we introduce an intrinsic reward mechanism that quantifies the trade-off between partial-layer and full-layer updates, incentivizing the model to achieve editing goals with minimal parameter modifications. Unlike existing approaches that perturb entire static layers (Figure~\ref{fig:open}(a)), HiEdit dynamically identifies and modifies only the most knowledge-relevant layers, maximizing editing efficiency while minimizing disruption.

Our main contributions are as follows:

\begin{itemize}
    \item We propose HiEdit, a novel framework for lifelong model editing, which utilizes hierarchical reinforcement learning to restructure the limited and complex action space of parameter updates into a structured hierarchical framework, achieving adaptive and localized knowledge updates.  
    \item We introduce an intrinsic reward mechanism based on the relative advantage of partial-layer versus full-layer updates, which encourages sparse and efficient layer selection, reducing unnecessary parameter modifications while preserving editing performance.
    \item To the best of our knowledge, we are the first to explore a challenging experimental setup involving timely and long-range sequential editing. Empirical results show that HiEdit improves the performance of the competitive RLEdit by an average of 8.48\% with perturbing only half of the layers per edit.
\end{itemize}

\section{Related Works}

\paragraph{Lifelong Model Editing (LME).} LME extends model editing to sequential scenarios, enabling hundreds to thousands of edits without compromising general performance or prior modifications. Traditional methods follow a ``locating-then-editing'' paradigm, identifying influential layers through causal tracing or search methods, then applying parameter perturbations. Approaches like GRACE \citep{hartvigsen2023aging}, RECT \citep{gu-etal-2024-model}, PRUNE \citep{ma2024perturbation}, AlphaEdit \citep{fang2024alphaedit}, and RLEdit \citep{li2025reinforced} offer various strategies for preserving knowledge during sequential editing. Unlike existing methods, HiEdit does not treat layer selection and layer update as isolated stages. Instead, it considers layer selection as a learnable high-level action, allowing for the dynamic  selection of LLM layers for updates based on varying knowledge. Detailed content is available in Appendix~\ref{app:rel}.

\paragraph{Hierarchical Reinforcement Learning (HRL).} HRL enhances exploration and learning efficiency through hierarchical structures, addressing complex action spaces. Methods are mainly categorized into option-based and subgoal-based approaches. Option-based HRL uses temporally extended actions, or options, to simplify decision-making through multi-step operations and task decomposition \citep{sutton1999between}, with studies focusing on autonomous option learning \citep{bacon2017option, harb2018waiting}. Subgoal-based HRL defines intermediate objectives to guide learning, traditionally constrained by manual sub-goals \citep{tessler2017deep}, but recent work enables autonomous subgoal discovery \citep{nachum2018data, jiang2025causal}. Our approach aligns with option-based HRL, decomposing layer selection and layer updating into subtasks managed by high-level and low-level hypernetworks, facilitating efficient exploration and learning. 

\section{Preliminary}

\subsection{Hypernetwork-based Model Editing}

HiEdit belongs to the hypernetwork-based model editing approaches, which involve training hyper-networks to generate parameter updates for language models. These methods utilize a set of small auxiliary editing networks $\{\mathcal{H}_1, \mathcal{H}_2,\dots,\mathcal{H}_L\}$ to transform gradients $\{\nabla_{\mathcal{W}_1}, \nabla_{\mathcal{W}_2},\dots,\nabla_{\mathcal{W}_L}\}$, obtained from standard fine-tuning, into parameter updates $\{\tilde{\nabla}_{\mathcal{W}_1}, \tilde{\nabla}_{\mathcal{W}_2},\dots,\tilde{\nabla}_{\mathcal{W}_L}\}$. Here, $L$ represents the size of the influential layer range identified prior to editing. By employing low-rank decomposition of gradients \citep{mitchell2021fast} and parameters \citep{hu2022lora}, the parameterization of this transformation becomes tractable. Specifically, each layer's gradient matrix is decomposed into a rank-1 product as $\nabla_{\mathcal{W}_l}=v_{l}u_l^{\mathsf{T}}$, where $l\in \{1,\dots,L\}$. Here, $u_l$ represents the inputs to layer $l$, and $v_l$ is the gradient of the standard fine-tuning loss with respect to the outputs of layer $l$. Through low-rank decomposition, each editing network $\mathcal{H}_l$ can efficiently learn a $d\to d$ mapping instead of the $d^2\to d^2$ mapping:

\begin{equation}
\hspace{-0.5cm}
\begin{aligned}
  \label{eq:1}
  \mathcal{H}_l: v_l\times u_l^\mathsf{T} \to \tilde{v}_l\times \tilde{u}_l^\mathsf{T},
\end{aligned}
\hspace{-0.5cm}
\end{equation}

where $\tilde{v}_l$ and $\tilde{u}_l$ are pseudo-vectors used to form the parameter updates as $\tilde{\nabla}_{\mathcal{W}_l}=\tilde{v}_l \tilde{u}_l^{\mathsf{T}}$. To further reduce the number of additional parameters, these methods share parameters across editor networks. They learn a separate set of editor parameters for each unique shape of the weight matrix to be edited and apply a layer-specific scale and offset module to the editor network's hidden states and outputs, enabling layer-wise specialization.

\subsection{RL for Lifelong Model Editing}

Lifelong model editing (LME) requires continuous knowledge updates on deployed LLMs, potentially reaching thousands or even tens of thousands of edits. In LME, a sequence of knowledge updates $[(x_1,y_1),(x_2,y_2),\dots,(x_T,y_T)]$ arrives in a streaming fashion, where $X=[x_1,x_2,\dots,x_T]$ is the is the input of knowledge to be edited, and $Y=[y_1,y_2,\dots.y_T]$ is the target output. Each knowledge pair $(x_t,y_t)$, where $t\in \{1,\dots,T\}$, comprising an input and its corresponding target output. The initially deployed LLM is defined as $f_{\mathcal{W}_0}:X\to Y'$ with parameters $\mathcal{W}_0$, mapping each input $x_t\in X$ to the original output $y'_t\in Y'$. At each step $t$ of sequential editing, the editor $\mathbf{ME}$ is tasked with modifying the model parameters to ensure $f_{\mathcal{W}_t}(x_t)=y_t$:

\begin{equation}
\hspace{-0.5cm}
\begin{aligned}
  \label{eq:2}
  f_{\mathcal{W}_t}=\mathbf{ME}(f_{\mathcal{W}_{t-1}},x_t,y_t).
\end{aligned}
\hspace{-0.5cm}
\end{equation}

Recent research \citep{li2025reinforced} frames LME as a reinforcement learning task, enabling the hypernetwork-based editor to accurately capture changes in LLMs and generate effective parameter updates for long-range sequential edits. Specifically, the hypernetwork training process in sequential editing can be modeled as a Markov Decision Process, where editing losses naturally serve as the immediate reward. Formally, at each time step $t$, the hypernetwork $\mathcal{H}_{\theta}$ with parameters $\theta$ generates an action $a_t$ for parameter updates $\tilde{\nabla}_{\mathcal{W}_{t}}$. The state $s_t$ consists of current LLM parameters $\mathcal{W}_{t-1}$ and the knowledge to edit $(x_{t}, y_{t})$. The transition function $\mathcal{P}$ deterministically updates the state as $s_{t+1} = \mathcal{P}(s_t,a_t)$. The immediate reward $r_t$ is derived from the negative of editing losses $\mathcal{L}_t$. After collecting all rewards along the entire editing trajectory, the policy parameterized by the hypernetwork $\mathcal{H}_{\theta}$ is optimized using the objective $J$:

\begin{equation}
\hspace{-0.5cm}
\begin{aligned}
  \label{eq:3}
  {\theta} = \underset{\hat{\theta}}{\arg\max}\ J = \underset{\hat{\theta}}{\arg\max}\sum\limits_{t=1}^T\gamma^t r_t,
\end{aligned}
\hspace{-0.5cm}
\end{equation}

where $T$ denotes the total length of the edit sequence, and $\gamma\in[0,1]$ is the discount factor. 

\begin{figure*}[ht]
  \centering
  \includegraphics[width=\linewidth]{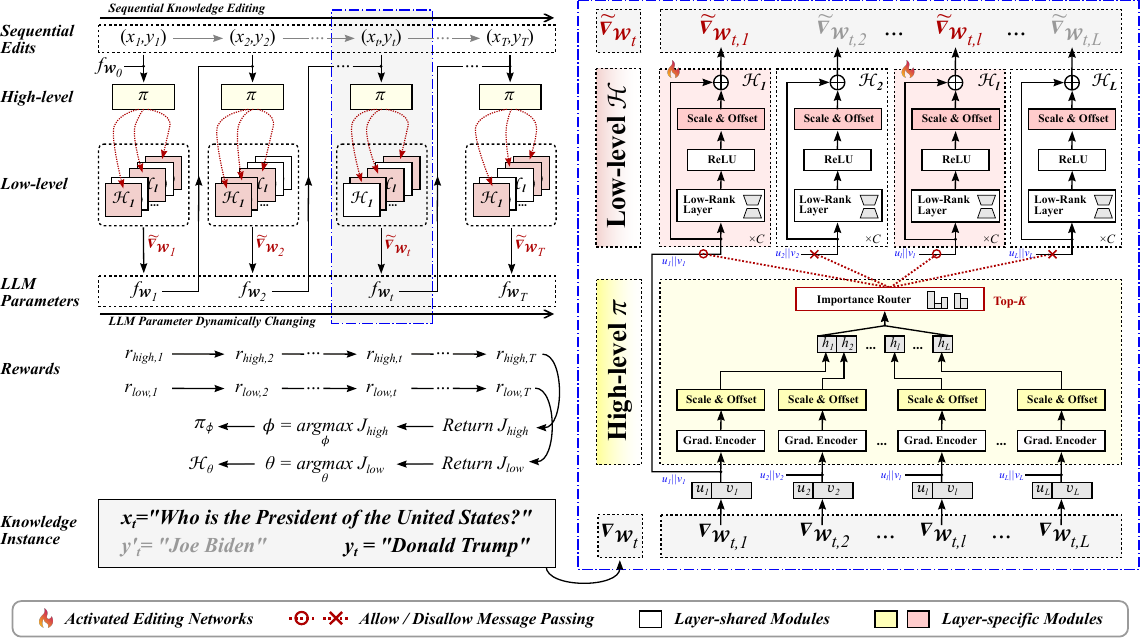}
  \caption {Illustration of lifelong model editing with HiEdit.
}
  \label{fig:main}
\end{figure*}

\section{Method}

In this section, we introduce HiEdit, a lifelong model editing approach using hierarchical reinforcement learning for adaptive and localized editing. Section~\ref{method:1} elaborates on the HRL paradigm for hypernetwork training within LME, including the Hierachical Markov Decision Process and the intrinsic reward mechanism. Section~\ref{method:2} details the design of model architectures for both high-level and low-level hypernetworks. Section~\ref{method:3} elucidates the training process of HiEdit.

\subsection{HRL for Lifelong Model Editing}
\label{method:1}

In Hierarchical Reinforcement Learning (HRL), the decision-making process is decomposed into two levels: high-level decisions, referred to as options, and low-level decisions, referred to as actions. This hierarchical structure effectively separates objectives across levels and reorganizes the complex action space, thereby enhancing the efficiency of exploration and policy learning.

Utilizing the hierarchical structure of decision-making, we formulate Lifelong Model Editing (LME) as a Hierarchical Markov Decision Process represented by the tuple $(\mathcal{S}, \mathcal{A}, \Omega, \mathcal{P}, r, \gamma)$, which consists of the state space $\mathcal{S}$, the action space $\mathcal{A}$, the option space $\Omega$, the transition function $\mathcal{P}$, the reward function $r$, and a discount factor $\gamma$. Given a sequence of knowledge updates $[(x_1, y_1),(x_2, y_2),\dots,(x_T, y_T)]$ and unrelated inputs $\tilde{X}=[\tilde{x}_1,\tilde{x}_2,\dots,\tilde{x}_T]$, at each time step $t$, the option $\omega_t \in\Omega$, where $\Omega=\{0,1\}^L$, represents the high-level action to select partial layers within the full influential layer range $\{1,\dots,L\}$. Subsequently, the low-level action $a_t\in\mathcal{A}$ produces parameter updates $\tilde{\nabla}_{\mathcal{W}_t}$ of the selected layers based on $\omega_t$. The transition function $\mathcal{P}$ deterministically updates the state $s_t\in\mathcal{S}$ as $\mathcal{P}(s_{t+1}|s_{t}, \omega_t, a_t)$, where $s_t$ comprises current LLM parameters $\mathcal{W}_{t-1}$ and the knowledge to edit $(x_t, y_t)$. We employ a high-level hypernetwork $\pi_{\phi}: \mathcal{S}\to\Omega$ and a low-level hypernetwork $\mathcal{H}_\theta:\mathcal{S}\times\Omega\to\mathcal{\mathcal{A}}$ to parameterize the high-level and low-level policies, respectively, with $\phi$ and $\theta$ denoting their respective parameters\footnote{Note that the $(s_t,\omega_t)$ pairs result in an augmented state space for the low-level hypernetwork $\mathcal{H}_\theta$.}. The reward signal is derived from the editing losses. Specifically, the immediate reward is decoupled into a high-level reward $r_{\text{high},t}$ for $\pi_\phi$ and a low-level reward $r_{\text{low},t}$ for $\mathcal{H}_\theta$. Following RLEdit \citep{li2025reinforced}, we define the low-level reward as the negative of the total loss: $r_{\text{low},t}=-\mathcal{L}_t$. Formally, the loss is expressed as $\mathcal{L}_t = \eta\|\tilde{\nabla}_{\mathcal{W}_t}\|^2 + \Sigma_{i=t-k}^{t}\mu^{t-i}\mathcal{L}_{t,i}$, where:

\begin{equation}
\hspace{-0.5cm}
\begin{aligned}
  \label{eq:4}
  \mathcal{L}_{t,i} & = - \log{p_{\mathcal{W}_t}(y_i|x_i)} + \tilde{\lambda}\tilde{\mathcal{L}}_{t,i}, \\
  \tilde{\mathcal{L}}_{t,i}& = \mathbf{KL}[p_{\mathcal{W}_{t-1}}(\cdot|\tilde{x}_{i})\|p_{\mathcal{W}_t}(\cdot|\tilde{x}_{i})].
\end{aligned}
\hspace{-0.5cm}
\end{equation}

Here, $\eta$ is the regularization coefficient, $\mu$ is the decay factor for memory backtracking, and $\tilde{\lambda}$ serves to balance 
the trade-off between updating target knowledge and preserving unrelated knowledge.

To promote sparse selection of editing-efficient layers that minimize impacts on other knowledge, we introduce an intrinsic reward mechanism. This mechanism measures the relative advantage of partial-layer versus full-layer updates, formulating the high-level reward $r_{\text{high},t}$ as an intrinsic reward:

\begin{equation}
\hspace{-0.5cm}
\begin{aligned}
  \label{eq:5}
  r_{\text{high},t} = r_{\text{low}}(s_t,\omega_t,a_t) - r_{\text{low}}(s_t,\mathbf{1},a_t),
\end{aligned}
\hspace{-0.5cm}
\end{equation}

where $\mathbf{1}=\{1\}^L$ denotes the selection of all layers within the influential layer range.

\subsection{Model Architecture of HiEdit}
\label{method:2}

As illustrated in Figure~\ref{fig:main}, HiEdit decouples the editing process into two distinct subtasks: layer selection and layer updating. These tasks are managed by the high-level hypernetwork $\pi_\phi$ and the low-level hypernetwork $\mathcal{H}_\theta$, respectively.

\subsubsection{The High-level Hypernetwork}

The high-level hypernetwork $\pi_{\phi}$ aims to generate a layer selection mask $m_t=[m_{t,1}, m_{t,2},\dots,m_{t,L}]$ base on the characteristics of the knowledge to edit $(x_t,y_t)$. This mask $m_t\in\{0,1\}^L$ serves as an option $\omega_t$ to determine which layers to edit. Specifically, if $m_{t,l}=1$, then the $l$-th layer is selected for editing, and the corresponding low-level editing network $\mathcal{H}_l$ is activated. 

At each time step $t$ of the the editing sequence, the gradient matrices of all influential parameters $\nabla_{\mathcal{W}_{t}}=\{\nabla_{\mathcal{W}_{t,1}},\nabla_{\mathcal{W}_{t,2}},\dots,\nabla_{\mathcal{W}_{t,L}}\}$ are obtained by standard fine-tuning for the knowledge pair $(x_t, y_t)$. Subsequently, each gradient matrix is decomposed using low-rank gradient decomposition as $\nabla_{\mathcal{W}_{t,l}}=v_l u_l^{\mathsf{T}}$. The decomposed vectors $u_l$ and $v_l$ are concatenated and fed into a layer-shared gradient encoder with weights $\mathbf{W}_{\text{GradEnc}}\in\mathbb{R}^{d_1\times d}$, where $d_1<d$ and $d_1,d$ are the dimensions of the weight matrix. A layer-specific scale and offset module $\mathbf{SPE}_l$ is then applied to the gradient encoder's hidden output to facilitate layer-wise specialization. Formally:

\begin{equation}
\hspace{-0.5cm}
\begin{aligned}
  \label{eq:6}
  h_l & = \mathbf{SPE}_l(\sigma(\mathbf{W}_{\text{GradEnc}}(u_l\|v_l))), \\
  z_t & = \mathbf{W}_{\text{GateNet}}(h_1\|h_2\|\dots\|h_L), \\
  m_t & = \mathbf{TopK}(z_t,K),
\end{aligned}
\hspace{-0.5cm}
\end{equation}

where $\sigma(\cdot)$ denotes the activation function (e.g., ReLU). The output features $h_l$ from each layer-specific module are concatenated and processed by a gate network with weights $\mathbf{W}_{\text{GateNet}}\in\mathbb{R}^{d_2\times L}$, where $d_2=d_1\cdot L$, yielding the layer importance distribution $z_t\in\mathbb{R}^L$. The mask $m_t$ is then generated from $z_t$ using $\mathbf{TopK}$. The indices corresponding to the top-$K$ largest values in $z_t$ are set to $1$ in $m_t$, while others are set to $0$. The gate network and the $\mathbf{TopK}$ module collectively form a critical component known as the importance router.

HiEdit distinguishes itself from existing methods by integrating layer selection and layer update into a unified, learnable hierarchical action space, rather than treating them as independent stages. Drawing inspiration from the straight-through estimator commonly used in Mixture of Experts (MoE) methods \citep{bengio2013estimating,tang2025graphmoe}, we employ a differentiable approximation to enhance effective gradient backpropagation for discrete routing within the importance router. Specifically, we employ a stopping gradient operator to decouple the forward and backward propagation processes while preserving equivalent output values. 

\begin{equation}
\hspace{-0.5cm}
\begin{aligned}
  \label{eq:7}
  m_t = \mathbf{sg}(m_t-z_t)+z_t,
\end{aligned}
\hspace{-0.5cm}
\end{equation}

where $\mathbf{sg}(\cdot)$ is the stop gradient operator which retains the forward output unchanged but sets the gradient to zero during backpropagation.

\subsubsection{The Low-level Hypernetwork}

The low-level hypernetwork $\mathcal{H}_\theta$ seeks to produce effective parameter updates $\tilde{\nabla}_{\mathcal{W}_t}$ from the gradient matrices\footnote{These gradient matrices only need to be computed once and then selectively fed into the activated low-level editing networks.} of partial layers selected by the high-level hypernetwork $\pi_\phi$. These parameter updates serve as an action $a_t$ to perturb current LLM parameters $\mathcal{W}_{t-1}$ as $\mathcal{W}_{t}=\mathcal{W}_{t-1}+\tilde{\nabla}_{\mathcal{W}_t}$. The low-level hypernetwork $\mathcal{H}_\theta$ comprises a set of small editing networks $\{\mathcal{H}_1,\mathcal{H}_2,\dots,\mathcal{H}_L\}$, but only partial editing networks are activated per edit. Specifically, only if $m_{t,l}=1$, the $l$-th editing network $\mathcal{H}_l$ is activated and transform the gradient matrix $\nabla_{\mathcal{W}_{t,l}}$ obtained by standard fine-tuning into parameter updates $\tilde{\nabla}_{\mathcal{W}_{t,l}}$. 

The implementation of $\mathcal{H}_\theta$ can follow either the MEND \citep{mitchell2021fast} or MALMEN \cite{tan2023massive} architecture. In the MEND-style implementation, each $\mathcal{H}_l$ comprises $C$ blocks, where the $c$-th block is parameterized by a shared linear layer with low-rank weight matrices $\mathbf{A}_{l}^{(c)}\in\mathbb{R}^{d_r\times d}$ and $\mathbf{B}_{l}^{(c)}\in\mathbb{R}^{d\times d_r}$, a residual connection, and a layer-specific scale and offset module $\mathbf{SPE}_{l}^{(c)}$. To edit layer $l$ at step $t$, the decomposed vectors $u_l$ and $v_l$ from the gradient matrix $\nabla_{\mathcal{W}_{t,l}}$ are concatenated and fed into the editing network $\mathcal{H}_l$, producing a final output of the same dimension.

\begin{equation}
\hspace{-0.5cm}
\begin{aligned}
  \label{eq:8}
  h_l^{(c)} = \mathbf{SPE}_l^{(c)}(\sigma(\mathbf{B}_l^{(c)}\cdot\mathbf{A}_l^{(c)}(h_l^{(c-1)})),
\end{aligned}
\hspace{-0.5cm}
\end{equation}

where $h_l^{(0)} = u_l\|v_l$ and $\sigma(\cdot)$ denotes the activation function (e.g., ReLU). The final output of the editing network $\mathcal{H}_l$ is split into pseudo vectors $\tilde{u}_l$ and $\tilde{v}_l$ as $h_l^{(C)}=\tilde{u}_l\|\tilde{v}_l$, ultimately yielding the parameter updates $\tilde{\nabla}_{\mathcal{W}_{t,l}}=\tilde{v}_l\tilde{u}_l^{\mathsf{T}}$.

\subsection{Training Process of HiEdit}
\label{method:3}

As shown in Figure~\ref{fig:main}, the parameters of the high-level hypernetwork and the activated low-level editing networks (marked with a flame icon) are jointly optimized after traversing the editing sequence and collecting rewards along the trajectory. By maintaining consistent sparsity during both training and inference, HiEdit encourages the hypernetworks to adaptively select sparse layers and generate appropriate parameter updates for specific knowledge. The accumulated high-level and low-level rewards are then utilized to update their respective hypernetworks:

\begin{equation}
\hspace{-0.5cm}
\begin{aligned}
  \label{eq:9}
  \alpha & = \underset{\hat{\alpha}}{\arg\max}\ J_{\beta} = \underset{\hat{\alpha}}{\arg\max}\ \sum\limits_{t=1}^T\gamma^t r_{\beta,t}, 
\end{aligned}
\hspace{-0.5cm}
\end{equation}

where $(\alpha,\beta)\in\{(\phi,\text{high}), (\theta,\text{low})\}$. The discount factor $\gamma$ is set to $1$ to ensure that the importance of all knowledge across the entire sequence is uniformly weighted during the training phrase.

\begin{table*}[ht]
  \centering
  \scalebox{0.8}{
  \begin{tabular}{c|c|cccc|cccc}
  \toprule
  \multirow{2}*{\textbf{Base Model}} & \multirow{2}*{\textbf{Editing Method}} & \multicolumn{4}{c}{\textsc{\textbf{ZsRE}}} & \multicolumn{4}{c}{\textsc{\textbf{CounterFact}}} \\
  \cmidrule(lr){3-6}\cmidrule(lr){7-10}
  & & \textit{Eff.}($\uparrow$) & \textit{Gen.}($\uparrow$) & \textit{Spe.}($\uparrow$) & \textit{Ret.}($\uparrow$) & \textit{Eff.}($\uparrow$) & \textit{Gen.}($\uparrow$) & \textit{Spe.}($\uparrow$) & \textit{Ret.}($\uparrow$) \\
  \midrule
  \multirow{13}*{\textsc{Llama-3-8B}} & Pre-edited & 36.58 & 35.89 & 38.64 & 38.26 & 7.20 & 9.10 & 89.77 & 8.70 \\
  \cmidrule(lr){2-10}
  & FT & 9.16 & 8.11 & 1.95 & 8.82 & 65.24 & \underline{57.99} & 44.03 & \textbf{62.80} \\
  & ROME & 3.29 & 3.24 & 0.66 & 2.67 & 59.26 & 56.45 & 48.52 & 48.30 \\
  & MEMIT & 0.00 & 0.00 & 0.12 & 0.00 & 18.61 & 16.94 & 16.93 & 16.20 \\
  & PRUNE & 0.00 & 0.00 & 0.00 & 0.00 & 0.00 & 0.00 & 0.00 & 0.00 \\
  & RECT & 0.00 & 0.00 & 0.00 & 0.00 & 0.00 & 0.00 & 0.00 & 0.00 \\
  & AlphaEdit & 16.22 & 14.58 & 3.87 & 9.10 & \textbf{73.06} & \textbf{66.43} & \underline{48.54}  & 47.40 \\
  & MEND & 0.00 & 0.00 & 0.00 & 0.02 & 7.69 & 13.40 & 7.45 & 10.50 \\
  & MALMEN & 5.09 & 4.74 & 0.72 & 4.26 & 0.00 & 0.00 & 0.00 & 0.00 \\
  & DAFNet & 21.15 & 20.41 & 21.15 & 21.13 & 29.56 & 31.55 & \textbf{68.38}& 33.60 \\
  & RLEdit & 81.43 & 79.49 & 42.73 & 70.72 & 66.35 & 55.26 & 44.79 & 57.20 \\
  & \cellcolor{gray!20}$\text{HiEdit}_\text{rand}$ & \cellcolor{gray!20}\underline{81.95} & \cellcolor{gray!20}\underline{79.63} & \cellcolor{gray!20}\underline{47.97} & \cellcolor{gray!20}\underline{74.66} & \cellcolor{gray!20}{66.40} & \cellcolor{gray!20}{55.48} & \cellcolor{gray!20}{45.51} & \cellcolor{gray!20}{58.00} \\
  & \cellcolor{gray!20}$\text{HiEdit}_\text{full}$ & \cellcolor{gray!20}\textbf{82.10} & \cellcolor{gray!20}\textbf{79.99} & \cellcolor{gray!20}\textbf{48.42} & \cellcolor{gray!20}\textbf{75.16} & \cellcolor{gray!20}\underline{66.53} & \cellcolor{gray!20}{55.65} & \cellcolor{gray!20}{45.70} & \cellcolor{gray!20}\underline{58.20} \\
  \bottomrule
  \toprule
  
  \multirow{13}*{\textsc{Gemma-2-9B}} & Pre-edited & 32.75 & 31.87 & 39.46 & 32.36 & 8.23 & 10.80 & 88.72 & 10.40 \\
  \cmidrule(lr){2-10}
  & FT & 35.38 & 32.31 & 30.46 & 33.40 & \underline{60.13} & 39.49 & 52.26 & 42.70 \\
  & ROME & 0.00 & 0.00 & 0.00 & 0.00 & 0.00 & 0.00 & 0.00 & 0.00 \\
  & MEMIT & 10.54 & 10.40 & 14.66 & 9.22 & 31.51  & 28.78  & 65.58  & 30.10 \\
  & PRUNE & 10.46  & 10.50  & 14.66  & 9.46 & 32.41  & 29.78  & 65.10  & 29.80 \\
  & RECT & 11.26 & 11.25 & 16.19 & 9.62 & 30.72 & 28.34 & 66.08 & 30.00 \\
  & AlphaEdit & 15.79 & 15.32 & 20.21 & 13.19 & 38.17 & 38.83 & 66.51 & 34.00 \\
  & MEND & 20.78 & 20.29 & 12.34 & 21.11 & 27.28 & 28.54 & 70.20 & 27.70 \\
  & MALMEN & 9.30 & 8.87 & 12.90 & 11.22 & 10.46 & 13.12 & \textbf{86.94} & 11.40 \\
  & DAFNet & 23.02 & 22.17 & 28.32 & 23.15 & 10.23 & 12.74 & \underline{86.60} & 13.50 \\
  & RLEdit & 69.11 & 65.87 & 25.17 & 57.66 & 55.51 & 49.92 & {46.28} & 45.80 \\
  & \cellcolor{gray!20}$\text{HiEdit}_\text{rand}$ & \cellcolor{gray!20}\textbf{82.65} & \cellcolor{gray!20}\textbf{78.98} & \cellcolor{gray!20}\underline{31.95} & \cellcolor{gray!20}\underline{67.89} & \cellcolor{gray!20}{59.81} & \cellcolor{gray!20}\underline{50.00} & \cellcolor{gray!20}{47.37} & \cellcolor{gray!20}\underline{50.30} \\
  & \cellcolor{gray!20}$\text{HiEdit}_\text{full}$ & \cellcolor{gray!20}\underline{82.12} & \cellcolor{gray!20}\underline{78.40} & \cellcolor{gray!20}\textbf{32.40} & \cellcolor{gray!20}\textbf{68.73} & \cellcolor{gray!20}\textbf{64.06} & \cellcolor{gray!20}\textbf{52.80} & \cellcolor{gray!20}43.59 & \cellcolor{gray!20}\textbf{54.00} \\
  \bottomrule
  \end{tabular}
  }
  \caption{
    Comparison of HiEdit with existing methods on the lifelong model editing (LME) task. \textit{Eff.}, \textit{Gen.}, \textit{Spe.}, and \textit{Ret.} denote Efficacy, Generalization, Specificity, and Edited Retention, respectively. The \textbf{Bold} and \underline{underline} mark the best and second-best results.
  }
  \label{tab:main}
\end{table*}

\section{Experiments}

We conduct extensive experiments to assess the effectiveness and scalability of HiEdit. Additionally, we perform a comprehensive ablation study to analyze the contribution of each component in HiEdit. To justify the computational efficiency of the additional high-level hypernetworks, we provide a detailed computational cost analysis. A case study visually demonstrates the editing effects of various methods on both previously and recently edited instances, highlighting the layers that HiEdit updates for these instances. Detailed results for the ablation study, case study, and computational cost analysis are included in Appendix~\ref{app:ab}, Appendix~\ref{app:co}, and Appendix~\ref{app:case}, respectively.

\subsection{Experimental Settings}

\paragraph{LLMs \& Datasets} We conduct experiments on two prominent auto-regressive LLMs: Llama-3-8B \citep{grattafiori2024llama} and Gemma-2-9B \citep{team2024gemma}. HiEdit and other baseline methods are evaluated using two widely utilized datasets for lifelong model editing: ZsRE \citep{levy-etal-2017-zero} and CounterFact \citep{meng2022locating}.

\paragraph{Evaluation Metrics.} Consistent with prior research \citep{meng2022locating,fang2024alphaedit,li2025reinforced}, we employ the metrics of Efficacy, Generalization, and Specificity to assess editing success. Additionally, we introduce a new metric named Edited Retention to evaluate the capability of lifelong model editing methods to retain knowledge from previous edits. This metric calculates the average of Efficacy and Generalization scores for the initially edited $T_0$ knowledge instances. More details are provided in Appendix~\ref{app:metrics}.

\paragraph{Baseline Methods.} We compare HiEdit against various model editing methods, including Fine-Tuning (FT) \citep{zhu2020modifying}, ROME \citep{meng2022locating}, MEMIT \citep{meng2022mass}, PRUNE \citep{ma2024perturbation}, RECT \citep{gu-etal-2024-model}, AlphaEdit \citep{fang2024alphaedit}, MEND \citep{mitchell2021fast}, MALMEN \citep{tan2023massive}, DAFNet \citep{zhang2024dafnet}, and RLEdit \citep{li2025reinforced}. More information is available in Appendix~\ref{app:baselines}.

\subsection{Main Results}
\label{exp:main}


In Table~\ref{tab:main}, we present a comprehensive comparison of HiEdit with various baseline methods on the lifelong model editing task. We randomly sample 8,000 knowledge instances from the ZsRE and CounterFact datasets, performing sequential editing with one knowledge instance per edit. This ``8000*1'' setup is designed for timely and long-range sequential editing, offering a more practical, dynamical and challenging scenario than the ``400*20'' and ``80*100'' setups utilized in previous research. The ``Pre-edited'' rows indicate the initial performance of LLMs before any editing is applied. The variants $\text{HiEdit}_{\text{full}}$ and $\text{HiEdit}_{\text{rand}}$ represent different approaches to intrinsic reward. Both RLEdit and HiEdit are implemented using two styles, with the results reflecting the style yielding optimal performance. Detailed comparisons of the intrinsic reward variations and implementation styles are discussed in Appendix~\ref{app:ab}.

Overall, HiEdit demonstrates superior performance across nearly all metrics, LLMs, and datasets. Compared to the most competitive baseline, RLEdit, HiEdit achieves average improvements of 9.02\%, 6.75\%, 11.60\%, and 11.28\% in Efficacy, Generalization, Specificity, and Edited Retention, respectively. These improvements are attributed to HiEdit's adaptive and localized parameter updates, which effectively minimize the impact on irrelevant and previously edited knowledge during sequential editing, while facilitating the integration of new knowledge. Other methods with high Specificity on the CounterFact dataset achieve this at the cost of significantly sacrificing Efficacy, Generalization, and Edited Retention performance, leading to severe editing failures.


\begin{table}[h]
  \centering
  \scalebox{0.8}{
  \begin{tabular}{c|cccc}
  \toprule
  \textbf{Editing Method} & \textit{Eff.}($\uparrow$) & \textit{Gen.}($\uparrow$) & \textit{Spe.}($\uparrow$) & \textit{Ret.}($\uparrow$) \\
  \midrule
  Pre-edited & 1.01 & 1.01 & 21.74 & 1.35 \\
  \midrule
  FT & 0.29 & 0.06 & 0.00 & 0.03 \\
  ROME & 1.78 & 1.80 & 1.17 & 0.90 \\
  AlphaEdit & 11.64 & 4.87 & 1.49 & 1.60 \\
  RLEdit & 27.93 & 20.58 & \underline{10.83} & 21.00 \\
  \rowcolor{gray!20}
  $\text{HiEdit}_{\text{rand}}$ & \underline{31.45} & \underline{22.35} & \textbf{10.96} & \underline{22.75} \\
  \rowcolor{gray!20}
  $\text{HiEdit}_{\text{full}}$ & \textbf{31.66} & \textbf{22.41} & 10.80 & \textbf{23.05} \\
  \bottomrule
  \end{tabular}
  }
  \caption{Comparison of HiEdit with competitive baselines on the CounterFact dataset and Llama-3-8B, employing more stringent top-1 accuracy metrics.}
  \label{tab:main2}
\end{table}

Notably, the CounterFact metrics discard redundant tokens from the original and target labels during computation to match their lengths, which can lead to insufficient and inaccurate evaluations when the target label is lengthy. In contrast, the ZsRE metrics directly evaluate the top-1 accuracy of all tokens in the target label, thereby offering a more stringent and accurate assessment. To more accurately assess HiEdit's performance on CounterFact, we employ the top-1 accuracy metrics to compare it with several competitive baselines. The results in Table~\ref{tab:main2} highlight HiEdit's ability to outperform other baseline methods.

It is observed that some methods yield results close to zero, which can be attributed to the challenging experimental setup involving timely and long-range sequential edits with thousands of consecutive parameter updates. Single-edit methods, such as ROME, MEMIT, MEND, and MALMEN, are specifically designed for isolated, one-time edits and struggle with knowledge conflicts and forgetting in sequential editing tasks. These limitations become even more pronounced in the timely and long-range experimental setup, explaining 0.0 results for these single-edit methods. In this challenging setup, LLM parameters undergo 8,000 sequential edits, with each edit introducing a new knowledge instance. Sequential-edit methods, such as PRUNE and RECT, often encounter issues like suboptimal editing effects, knowledge forgetting, or even model collapse, due to frequent updates to fixed and dense parameters. This explains 0.0 results for these sequential-edit methods.

\subsection{General Capabilities Tests}

We utilize six downstream tasks from the GLUE Benchmark \citep{wang2018glue} to assess the impact of various methods on the general capabilities of LLMs during sequential editing. Following prior studies \cite{fang2024alphaedit,li2025reinforced}, we evaluate the F1 score after sequentially editing Llama-3-8B using a configuration of 20,000 edits derived from the ZsRE dataset.

\begin{figure}[h]
  \centering
  \includegraphics[width=\linewidth]{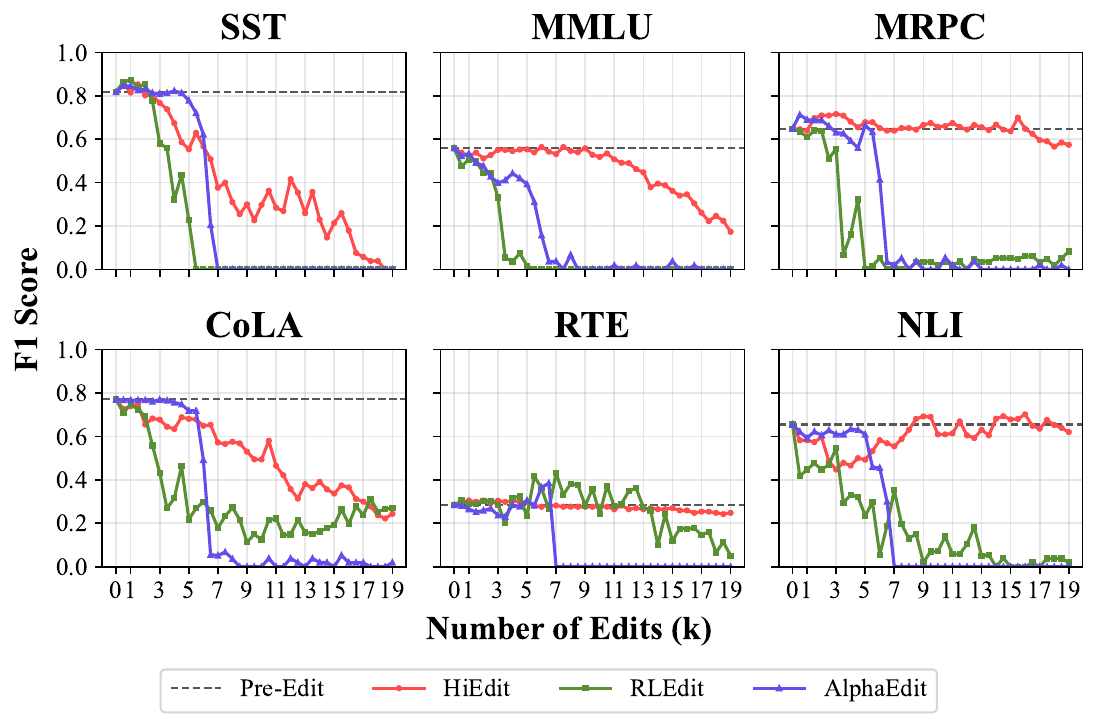}
  \caption {General capability assessment on six GLUE tasks during long-range sequential editing.}
  \label{fig:glue}
\end{figure}

As illustrated in Figure~\ref{fig:glue}, the performance of baseline methods on most general tasks experiences a sharp decline after 5,000 edits. In contrast, HiEdit demonstrates superior performance across all tasks. After 10,000 sequential edits, HiEdit's results on most tasks such as MMLU, MRPC, RTE, and NLI are comparable to those of the pre-edited model. Furthermore, HiEdit maintains strong performance even as the number of edits scales to 20,000, highlighting its effectiveness in preserving general capabilities during extensive sequential editing.

\subsection{Previously Edited Knowledge Tests}
\label{app:rent}

We assess the capability of various methods to retain previously edited knowledge during long-range sequential editing by evaluating the Efficacy, Generalization, and Specificity metrics for the initially edited $T_0=500$ knowledge instances on the ZsRE dataset and Llama-3-8B.

\begin{figure}[h]
  \centering
  \includegraphics[width=\linewidth]{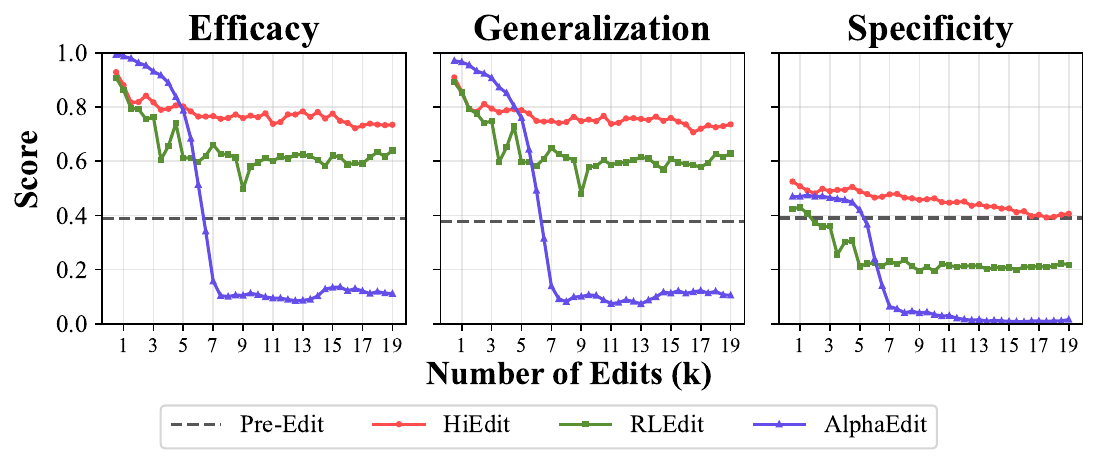}
  \caption {The metrics on the initially edited 500 knowledge instances during long-range sequential editing.}
  \label{fig:pre}
\end{figure}

As shown in Figure~\ref{fig:pre}, after 5,000 edits, all three metrics for the baseline methods exhibit a significant decline. In contrast, HiEdit maintains superior performance across all metrics, demonstrating enhanced robustness and stability as the number of edits increases. This highlights HiEdit's effectiveness in preserving previously edited knowledge during extensive sequential editing.

\subsection{Scaling Number of Edits Tests}

\begin{figure}[h]
  \centering
  \includegraphics[width=\linewidth]{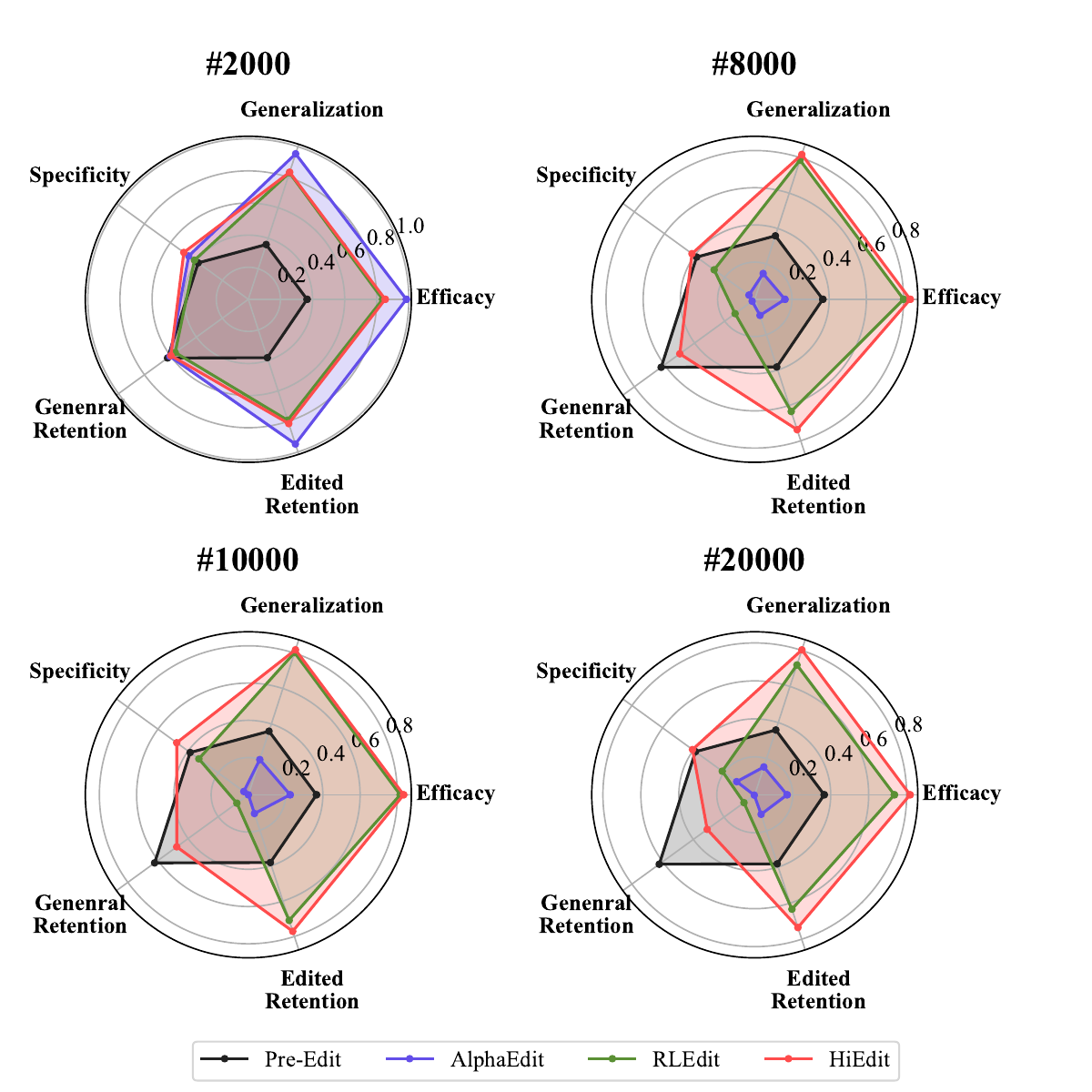}
  \caption {Performance comparison of vairous baseline methods across different number of edits.}
  \label{fig:radar}
\end{figure}

In Figure~\ref{fig:radar}, we compare the scalability of various methods as the number of sequential edits increases on Llama-3-8B, using configurations of 2,000, 8,000, 10,000, and 20,000 edits derived from the ZsRE dataset. Scalability is evaluated across five key metrics: Efficacy, Generalization, Specificity, General Retention, and Edited Retention. Notably, General Retention is calculated as the average F1 score across six general tasks from the GLUE Benchmark.

As illustrated in Figure~\ref{fig:radar}, AlphaEdit experiences a significant decline in performance at 8,000 edits, while RLEdit demonstrates better scalability due to its reinforcement learning framework. Leveraging adaptive and localized parameter updates, HiEdit consistently surpasses RLEdit across all metrics,  with particularly strong performance in Specificity, General Retention, and Edited Retention. Remarkably, HiEdit achieves General Retention scores comparable to the pre-edited model. Even at 20,000 sequential edits, HiEdit maintains strong performance across all metrics, underscoring its superior scalability and robustness in timely and long-range sequential editing scenarios.

\subsection{Hypotheses Tests}

To verify whether HiEdit identifies meaningful LLM layers for different knowledge samples, we conduct both quantitative and interpretative analyses on the ZsRE dataset and Llama-3-8B. 

\paragraph{Quantitative Comparison.} Using MEND-style hypernetworks, we compare HiEdit with two random selection baselines: (1) $\text{Random}_\text{1}$ randomly selects $K$ layers only during editing, using a well trained RLEdit hypernetwork; (2) $\text{Random}_\text{2}$: randomly selects $K$ layers during both hypernetwork training and editing. Table~\ref{tab:hypo} shows HiEdit consistently outperforms all random selection baselines, especially in Specificity and Edited retention, confirming its ability to identify meaningful layers for effective lifelong model editing.

\begin{figure}[h]
  \centering
  \includegraphics[width=\linewidth]{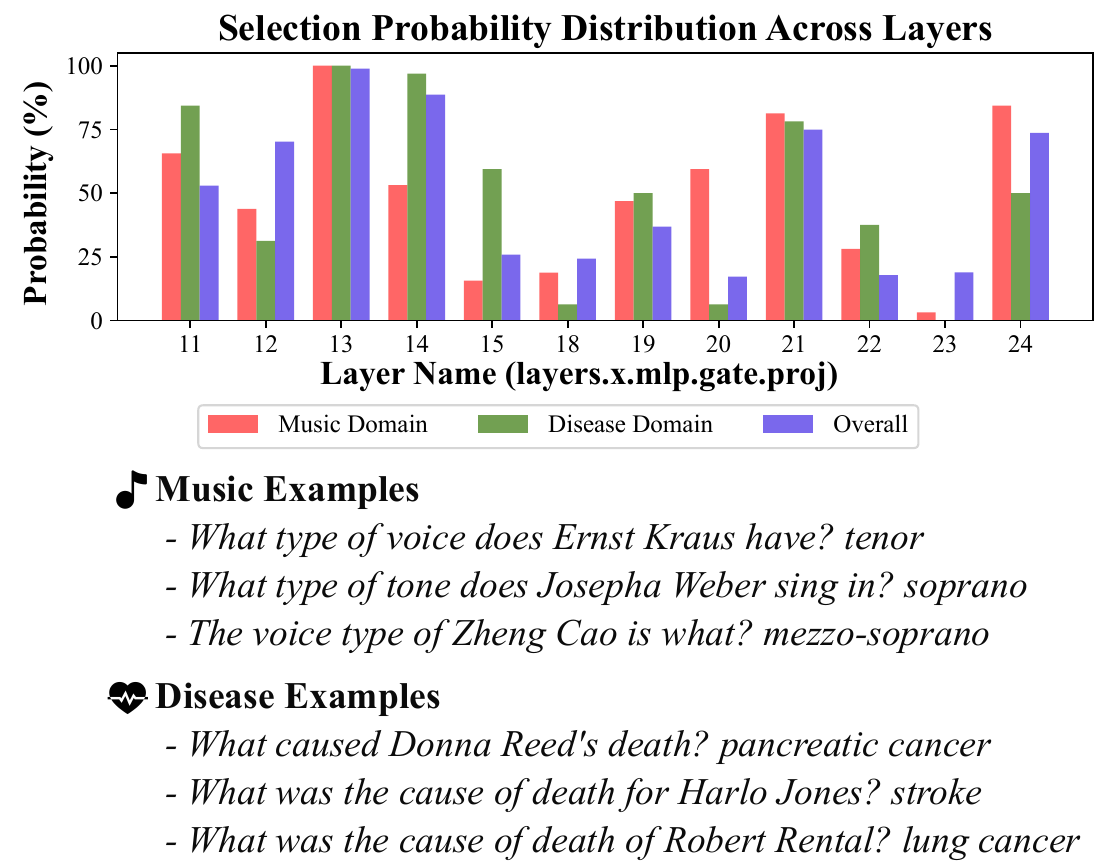}
  \caption {Visualization of layer selection pattern of HiEdit on ZsRE dataset and Llama-3-8B.}
  \label{fig:hypo}
\end{figure}

\paragraph{Intepretative Analysis.} We analyze HiEdit's layer selection pattern across 2,000 knowledge samples. As shown in Figure~\ref{fig:hypo}, the overall results (blue bars) indicate that certain layers (e.g., 13) are selected in over 88\% of samples, highlighting their central role in knowledge representation. Further analysis of 32 samples from two semantic category reveals domain-specific preferences: music knowledge (red bars) favors layers 20 and 24, while disease knowledge (green bars) tends to select layer 14 and 15. In Appendix~\ref{app:case}, we demonstrate the layers selected by HiEdit for different knowledge samples (highlighted in pink). These results indicate that HiEdit distinguishes between layers for various knowledge samples.

\begin{table}[h]
  \centering
  \scalebox{0.8}{
  \begin{tabular}{c|cccc}
  \toprule
  \textbf{Editing Method} & \textit{Eff.}($\uparrow$) & \textit{Gen.}($\uparrow$) & \textit{Spe.}($\uparrow$) & \textit{Ret.}($\uparrow$) \\
  \midrule
  Pre-edited  & 36.58 & 35.89 & 38.64 & 38.26 \\
  \midrule
  $\text{Random}_\text{1}$  & 81.00 & 79.42 & 30.32 & 69.72 \\
  $\text{Random}_\text{2}$ & 80.12 & 78.31 & 33.60 & 68.03 \\
  \rowcolor{gray!20}
  $\text{HiEdit}_{\text{rand}}$ & \underline{81.95} & \underline{79.63} & \underline{47.97} & \underline{74.66} \\
  \rowcolor{gray!20}
  $\text{HiEdit}_{\text{full}}$ & \textbf{82.10} & \textbf{79.99} & \textbf{48.42} & \textbf{75.16} \\
  \bottomrule
  \end{tabular}
  }
  \caption{Comparison of HiEdit with random selection of sparse layers for lifelong model editing on the ZsRE dataset and Llama-3-8B.}
  \label{tab:hypo}
\end{table}

\section{Conclusion}

We introduce HiEdit, a hierarchical reinforcement learning framework designed to adaptively select a minimal set of pertinent layers for each edit, thereby enhancing the integration of new knowledge while preserving existing information. Furthermore, we propose an intrinsic reward mechanism that measures the relative advantage of partial-layer versus full-layer updates, promoting minimal parameter modifications. HiEdit achieves fewer perturbations per edit, surpassing current state-of-the-art methods across various LLMs and datasets in target knowledge updates and existing knowledge preservation. It also exhibits enhanced scalability as the number of edits escalates.



\newpage

\section*{Limitations}

Our approach, HiEdit, has several limitations that warrant be acknowledged: (1) We conduct experiments on ZsRE and CounterFact datasets, which focus on modifying structured knowledge in general domains, but have not yet explored specific data domains or unstructured knowledge types; (2) We use the TopK mechanism to implement importance routing, where the number of editing layers $K$ is pre-fixed as a hyperparameter. Although effective, we have not yet explored dynamic mechanisms, such as Top-$p$, for more flexible selection of editing layers and other advanced implementations; (3) The potential for improved intrinsic reward design and reinforcement learning algorithms to enhance lifelong model editing performance has not been fully explored.


\section*{Acknowledgments}

We gratefully acknowledge the support of the National Natural Science Foundation of China (Grant No. 62350710797), the National Key Research and Development Program [2025YFE0209200], the Key Research and Development Program of Heilongjiang Province, China [2024ZX01A07], and the Science and Technology Innovation Award of Heilongjiang Province, China [JD2023GJ01].


\bibliography{custom}

\appendix

\section{Related Works}
\label{app:rel}

\paragraph{Lifelong Model Editing.} Lifelong Model Editing (LME) extends model editing in sequential scenarios, aiming to edit the deployed LLMs hundreds to thousands of times consecutively without compromising general performance or previous edits. Existing approaches typically follow a ``locating-then-editing'' paradigm. In this framework, the influential layer range is identified as a hyperparameter through causal tracing or search methods, and then parameter perturbations are applied to the entire static layer for knowledge editing. GRACE \citep{hartvigsen2023aging} dynamically updates an external codebook that maps old hidden states to new ones crucial for achieving the target output. RECT \citep{gu-etal-2024-model} employs regularization to preserve existing knowledge by constraining parameter shifts. PRUNE \citep{ma2024perturbation} minimizes disruption to existing information by controlling the singular values of the update matrix. AlphaEdit \citep{fang2024alphaedit} preserves knowledge stability by projecting parameter updates onto the null space of existing knowledge. RLEdit \citep{li2025reinforced} utilizes reinforcement learning to optimize hypernetwork parameters across entire editing trajectories, ensuring effective long-range sequential editing. Our proposed HiEdit establishes a structured hierarchical framework that expands the exploration space of RLEdit and enhances the exploration efficiency of hypernetworks. Unlike existing methods, HiEdit does not treat layer selection and layer update as isolated stages. Instead, it considers layer selection as a learnable high-level action, allowing for the dynamic selection of appropriate LLM layers for updates based on varying knowledge.

\paragraph{Hierarchical Reinforcement Learning.} Hierarchical Reinforcement Learning (HRL) employs hierarchical structures to enhance exploration and learning efficiency of agents, effectively addressing broad and complex action spaces. Methods are mainly categorized into option-based and subgoal-based approaches. Option-based HRL introduces temporally extended actions, or options, comprising a policy, termination condition, and initiation set \citep{sutton1999between}. This simplifies decision-making by enabling multi-step operations and task decomposition, facilitating efficient exploration and exploitation. Related studies \citep{bacon2017option, harb2018waiting} focus on auto-nomously learning options to enhance adaptability across diverse environments. Subgoal-based HRL defines intermediate objectives to guide learning, providing structured pathways toward long-range goals. Although traditionally constrained by manually specified sub-goals \citep{tessler2017deep}, related studies \citep{nachum2018data,jiang2025causal} enable agents to autonomously discover subgoals through state association, improving model flexibility and applicability. Our approach aligns with option-based HRL by introducing action abstractions for limited and complex parameter update actions, decoupling layer selection and layer updating into subtasks managed by high-level and low-level hypernetworks, facilitating efficient exploration and learning of the hypernetworks.

\section{Detailed Experimental Setup}

In this section, we detail our experimental setup, including five parts: datasets, metrics, baselines, GLUE benchmarks, and additional implementation detials.

\subsection{Datasets}
\label{app:datasets}

Here is the detailed introduction to the ZsRE and CounterFact datasets:

\paragraph{ZsRE.} Developed by \citep{levy-etal-2017-zero}, the ZsRE dataset is designed to assess models' capabilities in zero-shot relation extraction. Each entry in the dataset comprises a subject string and corresponding answers, which serve as targets for model editing. To evaluate generalization capabilities, the dataset includes questions rephrased through back-translation. Furthermore, unrelated locality questions are incorporated to examine models' specificity and its capacity to preserve unrelated knowledge. This dataset is pivotal for assessing success in efficacy, generalization, and specificity of model editing approaches. Following \citep{li2025reinforced}, we partition the ZsRE dataset into training and test sets, each containing approximately 20,000 knowledge instances.

\paragraph{CounterFact.} Introduced by \citep{meng2022locating}, the CounterFact dataset presents a more challenging benchmark that contrasts counterfactual statements with factual ones, thereby evaluating models' capabilities for managing contradictory information. The dataset constructs out-of-scope data by replacing the subject entity with approximate entities sharing the same predicate. Metrics similar to those used in ZsRE are applied to assess success in efficacy, generalization, and specificity of model editing methods. Following \citep{li2025reinforced}, we divide the CounterFact dataset into training and test sets, each comprising approximately 10,000 knowledge instances.

\subsection{Metrics}
\label{app:metrics}

\subsubsection{ZsRE Metrics}

Following prior research \citep{meng2022locating, meng2022mass, li2025reinforced}, we measures various model editing methods using standard metrics on the ZsRE dataset, calculating the average top-1 accuracy in the logits. Specifically, given an LLM $f_\mathcal{W}$, an editing knowledge pair $(x,y)$, equivalent knowledge $\bar{x}$, and unrelated knowledge pair $(\tilde{x}, \tilde{y})$, we assess the following metrics:

\paragraph{Efficacy.} This metric measures the success rate of editing the knowledge $(x,y)$ in $f_\mathcal{W}$. It involves comparing the top-1 logits output $\hat{y}=f_\mathcal{W}(x)$ with the target output $y$ when $x$ is input into $f_\mathcal{W}$:

\begin{equation}
\hspace{-0.5cm}
\begin{aligned}
  \label{eq:a1}
  \mathbb{E}\{y=\underset{\hat{y}}{\arg\max}{\ \mathbb{P}_{f_{\mathcal{W}}}(\hat{y}|x)}\}
\end{aligned}
\hspace{-0.5cm}
\end{equation}

\paragraph{Generalization.} This metric measures the success rate of editing equivalent knowledge $(\bar{x},y)$ in $f_\mathcal{W}$, which evaluates whether the LLM has effectively learned the intrinsic relationships within the knowledge and can extend to other equivalent knowledge. It involves comparing the top-1 logits output $\hat{y}=f_\mathcal{W}(\bar{x})$ with the target output $y$ when $\bar{x}$ is input into $f_\mathcal{W}$:

\begin{equation}
\hspace{-0.5cm}
\begin{aligned}
  \label{eq:a2}
  \mathbb{E}\{y=\underset{\hat{y}}{\arg\max}{\ \mathbb{P}_{f_{\mathcal{W}}}(\hat{y}|\bar{x})}\}
\end{aligned}
\hspace{-0.5cm}
\end{equation}

\paragraph{Specificity.} This metric measures the retention rate of unrelated knowledge $(\tilde{x}, \tilde{y})$ after editing, which evaluates whether the knowledge editing maintains locality and only modifies the target knowledge. It involves comparing the top-1 logits output $\hat{y}=f_\mathcal{W}(\tilde{x})$ with the original output $\tilde{y}$ when $\tilde{x}$ is input into $f_\mathcal{W}$:

\begin{equation}
\hspace{-0.5cm}
\begin{aligned}
  \label{eq:a3}
  \mathbb{E}\{\tilde{y}=\underset{\hat{y}}{\arg\max}{\ \mathbb{P}_{f_{\mathcal{W}}}(\hat{y}|\tilde{x})}\}
\end{aligned}
\hspace{-0.5cm}
\end{equation}

\subsubsection{CounterFact Metrics}

Similarly, following prior research \citep{meng2022locating, meng2022mass, li2025reinforced}, we measures various model editing methods using standard metrics on the CounterFact dataset, comparing the probabilities of different answers in the logits. Specifically, given an LLM $f_\mathcal{W}$, an editing knowledge pair $(x,y)$, original output $y'$, equivalent knowledge $\bar{x}$, and unrelated knowledge pair $(\tilde{x}, \tilde{y})$, we assess the following metrics:

\paragraph{Efficacy.} This metric measures the success rate of editing the knowledge $(x,y)$ in $f_\mathcal{W}$. It involves comparing whether the probability of the target output $y$ is higher than of the original output $y'$ in the logits when $x$ is input into $f_\mathcal{W}$:

\begin{equation}
\hspace{-0.5cm}
\begin{aligned}
  \label{eq:a4}
  \mathbb{E}[\mathbb{P}_{f_{\mathcal{W}}}(y|x) > \mathbb{P}_{f_{\mathcal{W}}}(y'|x)]
\end{aligned}
\hspace{-0.5cm}
\end{equation}

\paragraph{Generalization.} This metric measures the success rate of editing equivalent knowledge $(\bar{x},y)$ in $f_\mathcal{W}$, which evaluates whether the LLM has effectively learned the intrinsic relationships within the knowledge and can extend to other equivalent knowledge. It involves comparing whether the probability of the target output $y$ is higher than of the original output $y'$ in the logits when $\bar{x}$ is input into $f_\mathcal{W}$:

\begin{equation}
\hspace{-0.5cm}
\begin{aligned}
  \label{eq:a5}
  \mathbb{E}[\mathbb{P}_{f_{\mathcal{W}}}(y|\bar{x}) > \mathbb{P}_{f_{\mathcal{W}}}(y'|\bar{x})]
\end{aligned}
\hspace{-0.5cm}
\end{equation}

\paragraph{Specificity.} This metric measures the retention rate of unrelated knowledge $(\tilde{x}, \tilde{y})$ after editing, which evaluates whether the knowledge editing maintains locality and only modifies the target knowledge. It involves comparing whether the probability of the original output $\tilde{y}$ is higher than of the edited output $y$ in the logits when $\tilde{x}$ is input into $f_\mathcal{W}$:

\begin{equation}
\hspace{-0.5cm}
\begin{aligned}
  \label{eq:a6}
  \mathbb{E}[\mathbb{P}_{f_{\mathcal{W}}}(\tilde{x}|\tilde{x}) > \mathbb{P}_{f_{\mathcal{W}}}(y|\tilde{x})]
\end{aligned}
\hspace{-0.5cm}
\end{equation}

\subsection{Baselines}
\label{app:baselines}

We employed the code from AlphaEdit \citep{fang2024alphaedit} and RLEdit \citep{li2025reinforced} to assess the performance of baseline methods. Here is the detailed introduction to the baseline methods:

\paragraph{FT.} Fine-Tuning (FT) \citep{zhu2020modifying} is a traditional approach that directly updates the model parameters using standard gradient descent. Specifically, it employs an autoregressive loss function on the new knowledge to fine-tune specific layers of the LLM, typically the final few layers, to integrate the edit while attempting to minimize deviation from the original weights.

\paragraph{ROME.} Rank-One Model Editing (ROME) \citep{meng2022locating} is a locate-then-edit method designed to modify specific factual associations. It first utilizes causal tracing to identify the specific feed-forward neurons in the middle layers that are responsible for mediating factual knowledge. Subsequently, it treats the weight update as a rank-one modification problem, solving for the optimal update using Lagrange multipliers to insert the new fact while preserving existing knowledge.

\paragraph{MEMIT.} Mass-Editing Memory in a Transformer (MEMIT) \citep{meng2022mass} extends the principles of ROME to the mass-editing setting. Instead of updating a single layer for a single fact, MEMIT distributes the information storage across multiple MLP layers. It formulates the parameter update as a least-squares problem, allowing for the simultaneous insertion of thousands of factual associations into the model without significant performance degradation.

\paragraph{PRUNE.} PRUNE \citep{ma2024perturbation} addresses the challenges of sequential model editing by focusing on preserving the model's general capabilities. It introduces a condition number constraint on the parameter update matrix. By limiting the sensitivity of the edited parameters and controlling the singular values of the update matrix, PRUNE restricts the interference of new edits on previously stored knowledge, thereby mitigating the risk of model collapse during continuous updates.

\paragraph{RECT.} Regularization-based Editing (RECT) \citep{gu-etal-2024-model} is designed to alleviate the "catastrophic forgetting" of general reasoning abilities often caused by sequential editing. It incorporates a regularization term into the optimization objective that constrains the magnitude of weight updates. By preventing the parameters from drifting excessively during the editing process, RECT aims to balance editing success with the preservation of the LLM's fundamental capabilities.

\paragraph{AlphaEdit.} AlphaEdit \citep{fang2024alphaedit} is a sequential editing method that leverages the geometric properties of the parameter space. It projects parameter updates onto the null space of the covariance matrix of previously learned knowledge. This projection ensures that new edits are orthogonal to, and therefore do not interfere with, the features required to recall existing knowledge, effectively mitigating the interference between consecutive updates in a lifelong editing scenario.

\paragraph{MEND.} Model Editor Networks with Gradient Decomposition (MEND) \citep{mitchell2021fast} represents a hypernetwork-based approach. Instead of directly optimizing the model parameters, MEND trains a hypernetwork to map the gradients obtained from standard fine-tuning into effective parameter updates. It utilizes a low-rank decomposition of the gradients to make this process computationally efficient, enabling fast and localized edits.

\paragraph{MALMEN.} Mass-Editing Language Models via Meta-Learning (MALMEN) \citep{tan2023massive} adapts the hypernetwork architecture for massive editing tasks. It aggregates the parameter shifts required for a large batch of edits by solving normal equations within a least-squares framework. This formulation allows the hypernetwork to generate a unified update that accounts for conflicts within the batch, separating the computation into memory-efficient steps suitable for large-scale updates.

\paragraph{DAFNet.} Dynamic Auxiliary Fusion Network (DAFNet) \citep{zhang2024dafnet} is tailored for sequential editing by enhancing standard hypernetworks with an auxiliary fusion module. This module captures the semantic interactions and context within the sequence of knowledge triples. By dynamically fusing this auxiliary information with the edit requests, DAFNet improves the model's ability to rectify mistakes continuously and adapt to evolving knowledge streams.

\paragraph{RLEdit.} Reinforced Lifelong Editing (RLEdit) \citep{li2025reinforced} formulates the hypernetwork training process as a Reinforcement Learning (RL) task. It models the lifelong editing process as a Markov Decision Process (MDP), where the hypernetwork acts as an agent generating updates (actions) to maximize a cumulative reward defined by editing efficacy and stability. RLEdit employs an offline policy update strategy and incorporates a memory backtracking mechanism to review previous edits, ensuring robustness and stability over long sequences of edits.

\subsection{GLUE Benchmarks}

The GLUE (General Language Understanding Evaluation) benchmark, developed by \citep{wang2018glue}, is a comprehensive suite of resources for training, evaluating, and analyzing natural language understanding systems. Following \citep{fang2024alphaedit, li2025reinforced}, we selected 6 tasks from this benchmark to evaluate the ability of various model editing methods to maintain general language capabilities:

\paragraph{SST (The Stanford Sentiment Treebank).} Introduced by \citep{socher2013recursive}, this dataset comprises movie review sentences annotated with sentiment labels. The binary classification task requires models to determine the sentiment expressed in each sentence.

\paragraph{MRPC (Microsoft Research Paraphrase Corpus).} As described by \citep{dolan2005automatically}, this benchmark assesses semantic similarity, challenging models to determine whether two sentences are semantically equivalent.

\paragraph{MMLU (Massive Multi-task Language Understanding).} Developed by \citep{hendrycks2021measuring}, this robust benchmark aims to evaluate language models across multiple domains, with a particular focus on zero-shot and few-shot settings.

\paragraph{RTE (Recognizing Textual Entailment).} Explored by \citep{bentivogli2009fifth}, this task involves analyzing logical relationships between sentences, requiring models to determine if a premise sentence logically entails a hypothesis sentence.

\paragraph{CoLA (Corpus of Linguistic Acceptability).} Introduced by \citep{warstadt2019neural}, this single-sentence classification task focuses on grammatical judgment, requiring models to differentiate between grammatically acceptable and unacceptable sentences extracted from linguistic literature.

\paragraph{NLI (Natural Language Inference).} Evaluated by \citep{williams2018broad}, this task assesses natural language understanding by requiring models to analyze pairs of sentences and determine their logical relationships.

\subsection{Implementation Details}

Our most experiments are conducted on a single NVIDIA-A100-80GB GPU. The LLMs are loaded using the HuggingFace Transformers library \citep{wolf2020transformers} and operated in half-precision mode to ensure a fair comparison. The hyperparameter configurations for HiEdit are summarized in Table~\ref{tab:hyperparam}, while other experimental configurations remain consistent with RLEdit \citep{li2025reinforced}.

\begin{table}[h]
  \centering
  \scalebox{0.7}{
  \begin{tabular}{ccccc}
  \toprule
  \textbf{Dataset} & \textbf{Model} & \textbf{Influential Layer Range} & \textbf{$K$} & \textbf{$d_1$} \\
  \midrule
  \multirow{2}*{{ZsRE}} & {Llama-3-8B} & gate[11-15],up[18-24] & 6 & 256 \\
   & {Gemma-2-9B} & gate[32-40],up[32-40] & 9 & 256 \\
   \midrule
  \multirow{2}*{{Counterfact}} & {Llama-3-8B} & gate[22-30],up[22.30] & 9 & 256 \\
   & {Gemma-2-9B} & gate[32-40],up[32-40] & 9 & 256 \\
  \bottomrule
  \end{tabular}
  }
  \caption{Hyperparameter configurations of HiEdit.}
  \label{tab:hyperparam}
\end{table}

The hyperparameter ``influential layer range'' is chosen empirically based on strong experimental performance reported in RLEdit \citep{li2025reinforced}. To ensure fairness in comparison, we adopt these ranges for HiEdit while introducing an additional hyperparameter, $K$, to limit the number of editing layers. For rigorous evaluation, we set $K$ as half of the ``influential layer range'' size $L$, i.e., $K=L/2$. In practice, $K$ can be flexibly chosen between $L/2$ and $L$ to optimize lifelong model editing performance. Results presented in Figure~\ref{fig:n_layers} demonstrate that limiting $K$ to values between $L/2$ and $L$ improves editing success (Efficacy and Generalization) while reducing editing impact (Specificity and General Retention) compared to editing all influential layers, as done in RLEdit.

\begin{table*}[ht]
  \centering
  \scalebox{0.75}{
  \begin{tabular}{c|c|c|cccc|cccc}
  \toprule
  \multirow{2}*{\textbf{Model}} & \multirow{2}*{\textbf{HyperNet}} & \multirow{2}*{\textbf{Method}} & \multicolumn{4}{c}{\textsc{\textbf{ZsRE}}} & \multicolumn{4}{c}{\textsc{\textbf{CounterFact}}} \\
  \cmidrule(lr){4-7}\cmidrule(lr){8-11}
  & & & \textit{Eff.}($\uparrow$) & \textit{Gen.}($\uparrow$) & \textit{Spe.}($\uparrow$) & \textit{Ret.}($\uparrow$) & \textit{Eff.}($\uparrow$) & \textit{Gen.}($\uparrow$) & \textit{Spe.}($\uparrow$) & \textit{Ret.}($\uparrow$) \\
  \midrule
  \multirow{12}*{{\textsc{Llama-3-8B}}} & \multirow{6}*{MEND} & \cellcolor{gray!20}$\text{HiEdit}_{\text{full}}$ & \cellcolor{gray!20}\textbf{83.58} & \cellcolor{gray!20}\textbf{81.88} & \cellcolor{gray!20}\underline{41.75} & \cellcolor{gray!20}\underline{73.73} & \cellcolor{gray!20}\textbf{66.53} & \cellcolor{gray!20}\textbf{55.65} & \cellcolor{gray!20}45.70 & \cellcolor{gray!20}\textbf{58.20} \\
   & & \cellcolor{gray!20}$\text{HiEdit}_{\text{rand}}$ & \cellcolor{gray!20}\underline{82.99} & \cellcolor{gray!20}\underline{81.15} & \cellcolor{gray!20}\textbf{43.15} & \cellcolor{gray!20}\textbf{74.06} & \cellcolor{gray!20}66.40 & \cellcolor{gray!20}\underline{55.48} & \cellcolor{gray!20}45.51 & \cellcolor{gray!20}58.00 \\
   & & w/o Advantage & {80.32} & {79.12} & 37.60 & 70.01 & \underline{66.41} & {55.34} & 44.85 & \underline{58.10} \\ 
   & & w/o RL training & 36.84 & 36.23 & {39.01} & 37.82 & 8.97 & 11.15 & \textbf{87.93} & 10.50 \\
   & & w/o HiNet (RLEdit) & 80.06 & 78.72 & 27.00 & 63.43 & 66.35 & 55.26 & 44.79 & 57.20 \\
   & & HiNet $\to$ Grad. & 74.56 & 72.82 & 23.32 & 58.94 & 65.69 & 54.33 & \underline{46.41} & 54.90 \\

  \cmidrule(lr){2-11}
   & \multirow{6}*{MALMEN} & \cellcolor{gray!20}$\text{HiEdit}_{\text{full}}$ & \cellcolor{gray!20}\underline{82.10} & \cellcolor{gray!20}\underline{79.99} & \cellcolor{gray!20}\textbf{48.42} & \cellcolor{gray!20}\textbf{75.16} & \cellcolor{gray!20}\textbf{61.96} & \cellcolor{gray!20}\textbf{51.24} & \cellcolor{gray!20}48.63 & \cellcolor{gray!20}\textbf{56.60} \\
   & & \cellcolor{gray!20}$\text{HiEdit}_{\text{rand}}$ & \cellcolor{gray!20}81.95 & \cellcolor{gray!20}79.63 & \cellcolor{gray!20}\underline{47.97} & \cellcolor{gray!20}74.66 & \cellcolor{gray!20}\underline{61.36} & \cellcolor{gray!20}50.55 & \cellcolor{gray!20}{49.18} & \cellcolor{gray!20}\underline{56.00} \\
   & & w/o Advantage & \textbf{83.30} & \textbf{81.35} & {47.85} & \underline{75.10} & {61.33} & \underline{50.70} & 49.40 & 54.90 \\ 
   & & w/o RL training & 37.61 & 36.70 & 38.82 & 43.62 & 9.82 & 11.80 & \textbf{87.23} & 13.90 \\
   & & w/o HiNet (RLEdit) & 81.43 & 79.49 & 42.73 & 70.72 & 58.60 & 49.64 & 50.69 & 52.80 \\
   & & HiNet $\to$ Grad. & 75.32 & 73.39 & 38.92 & 65.38 & 55.96 & 47.98 & \underline{52.46} & 50.60 \\

  \bottomrule
  \toprule
  \multirow{12}*{{\textsc{Gemma-2-9B}}} & \multirow{6}*{MEND} & \cellcolor{gray!20}$\text{HiEdit}_{\text{full}}$ & \cellcolor{gray!20}\underline{82.12} & \cellcolor{gray!20}{78.40} & \cellcolor{gray!20}\underline{32.40} & \cellcolor{gray!20}\textbf{68.73} & \cellcolor{gray!20}\textbf{64.06} & \cellcolor{gray!20}\textbf{52.80} & \cellcolor{gray!20}43.59 & \cellcolor{gray!20}\textbf{54.00} \\
   & & \cellcolor{gray!20}$\text{HiEdit}_{\text{rand}}$ & \cellcolor{gray!20}\textbf{82.65} & \cellcolor{gray!20}\textbf{78.98} & \cellcolor{gray!20}31.95 & \cellcolor{gray!20}67.89 & \cellcolor{gray!20}\underline{59.81} & \cellcolor{gray!20}\underline{50.00} & \cellcolor{gray!20}\underline{47.37} & \cellcolor{gray!20}\underline{50.30} \\
   & & w/o Advantage & {81.53} & \underline{78.50} & 30.41 & \underline{68.52} & {57.93} & 49.30 & {46.95} & 49.40 \\ 
   & & w/o RL training & 33.72 & 32.82 & \textbf{39.79} & 32.78 & 11.81 & 14.22 & \textbf{84.76} & 13.30 \\
   & & w/o HiNet (RLEdit) & 69.11 & 65.87 & 25.17 & 57.66 & 55.51 & {49.92} & 46.28 & 45.80 \\
   & & HiNet $\to$ Grad. & 66.32 & 63.43 & 23.81 & 52.19 & 55.65 & 48.46 & 46.50 & 43.00 \\

   \cmidrule(lr){2-11}
  & \multirow{6}*{MALMEN} & \cellcolor{gray!20}$\text{HiEdit}_{\text{full}}$ & \cellcolor{gray!20}\textbf{67.97} & \cellcolor{gray!20}\textbf{65.99} & \cellcolor{gray!20}\underline{37.63} & \cellcolor{gray!20}\underline{63.32} & \cellcolor{gray!20}\textbf{55.27} & \cellcolor{gray!20}\textbf{49.25} & \cellcolor{gray!20}46.48 & \cellcolor{gray!20}\underline{46.20} \\
   & & \cellcolor{gray!20}$\text{HiEdit}_{\text{rand}}$ & \cellcolor{gray!20}\underline{67.65} & \cellcolor{gray!20}\underline{65.66} & \cellcolor{gray!20}36.08 & \cellcolor{gray!20}\textbf{64.33} & \cellcolor{gray!20}\underline{55.03} & \cellcolor{gray!20}\underline{48.32} & \cellcolor{gray!20}46.70 & \cellcolor{gray!20}\textbf{47.30} \\
   & & w/o Advantage & 58.34 & 56.80 & 30.27 & 53.77 & {51.10} & {44.77} & 48.55 & 42.60 \\ 
   & & w/o RL training & 32.93 & 32.06 & \textbf{39.58} & 32.38 & 34.35 & 33.56 & \textbf{62.85} & 31.90 \\
   & & w/o HiNet (RLEdit) & {62.23} & {60.10} & 27.47 & 54.32 & 46.98 & 43.11 & 52.02 & 40.20 \\
   & & HiNet $\to$ Grad. & 59.87 & 58.00 & 29.52 & 51.45 & 41.52 & 38.36 & \underline{57.26} & 36.00 \\

  \bottomrule
  \end{tabular}
  }
  \caption{
    Ablation Study Results for HiEdit. The \textbf{Bold} and \underline{underline} mark the best and second-best results. 
  }
  \label{tab:ablation}
\end{table*}

\section{Ablation Study}
\label{app:ab}

To assess the contribution of each component in HiEdit, we conduct extensive ablation studies in Table~\ref{tab:ablation} using the same setup as in Section~\ref{exp:main}. This ablation study includes: the intrinsic reward mechanism, the hierarchical design, the hierarchical reinforcement learning (HRL) training strategy, the implementation styles, and the number of editing layers.

\subsection{The Intrinsic Reward Mechanism}

We conduct an experiment by removing the relative advantage from $\text{HiEdit}_\text{full}$, which is calculated through partial-layer and full-layer updates. In ``w/o Advantage'', the rewards for both high-level and low-level hypernetworks are equal, with $r_{\text{high}}=r_{\text{low}}=-\mathcal{L}$. Compared to $\text{HiEdit}_\text{full}$, the removal results in an average performance decrease of 4.03\%. This underscores the critical importance of our proposed relative advantage-based intrinsic reward mechanism in enhancing the lifelong model editing performance.

Additionally, we implemented a variant of HiEdit, ``$\text{HiEdit}_\text{rand}$'' by modifying the minuend in the relative advantage formula from the full-layer updates to a randomly selected $K$ layers to update. Results show that ``$\text{HiEdit}_\text{rand}$'' experiences an average performance decrease of 0.62\% compared to ``$\text{HiEdit}_\text{full}$''. Despite this decrease, ``$\text{HiEdit}_\text{rand}$'' still outperforms the removal implementation ``w/o Advantage'' by an average performance improvement of 3.82\%. This demonstrates that relative advantage provides a robust intrinsic reward mechanism, and full-layer updates serve as both an intuitive and efficient computational baseline.

\subsection{The Hierarchical Design}

Here, we conduct ``w/o HiNet'' by eliminating the high-level hypernetwork from $\text{HiEdit}_\text{full}$, resulting in its degeneration into RLEdit. Compared to $\text{HiEdit}_\text{full}$, the average performance decreased by 8.09\%. This indicates that limited and flat action space of parameter updates is not efficient enough for exploration. Our proposed hierarchical design introduces a structured action space, enhancing the exploration efficiency of the hypernetworks, thereby improving the lifelong model editing performance.

In ``$\text{HiNet}\to\text{Grad.}$'', we modify the layer selection method based on high-level hypernetwork exploration to a heuristic gradient-guided method, where the $K$ largest matrices, determined by the Frobenius norm obtained from standard fine-tuning, are selected for updating. Relative to $\text{HiEdit}_\text{full}$, the average performance decreased by 11.95\%. This suggests that relying solely on gradient norms is insufficient for effective layer selection. Instead, optimal layer selection should be learned from comprehensive gradient information, emphasizing the significance of hierarchical reinforcement learning modeling.

\subsection{The HRL Training Strategy}

Note that \citep{li2025reinforced} has already demonstrated the benefits of the reinforcement learning training strategy for low-level hypernetworks; thus, our focus here is on the training strategy of high-level hypernetworks. In ``w/o RL training'', we optimize the high-level hypernetwork parameters immediately after each edit, rather than cumulatively optimizing them after traversing the entire editing trajectory. The results indicate that, compared to $\text{HiEdit}_\text{full}$, the average performance decreased by 41.41\%. These findings highlight the pivotal role of the hierarchical reinforcement learning training strategy in enhancing the lifelong model editing performance.

\subsection{The Implementation Styles}

Table~\ref{tab:ablation} presents a comprehensive comparison of the lifelong model editing performance of RLEdit and HiEdit using two implementation styles for low-level hypernetworks: MEND and MALMEN. The results demonstrate that in the context of lifelong model editing, a challenging setting characterized by timely, long-range edits, the MEND style of RLEdit enhances average performance by 2.29\% compared to the MALMEN style. Similarly, the MEND style of HiEdit enhances average performance by 4.99\% relative to the MALMEN style. These findings suggest that the MEND style implementation may be better suited for this kind of timely and long-range lifelong model editing.

\subsection{The Number of Editing Layers}
\label{app:k}

\begin{figure}[h]
  \centering
  \includegraphics[width=\linewidth]{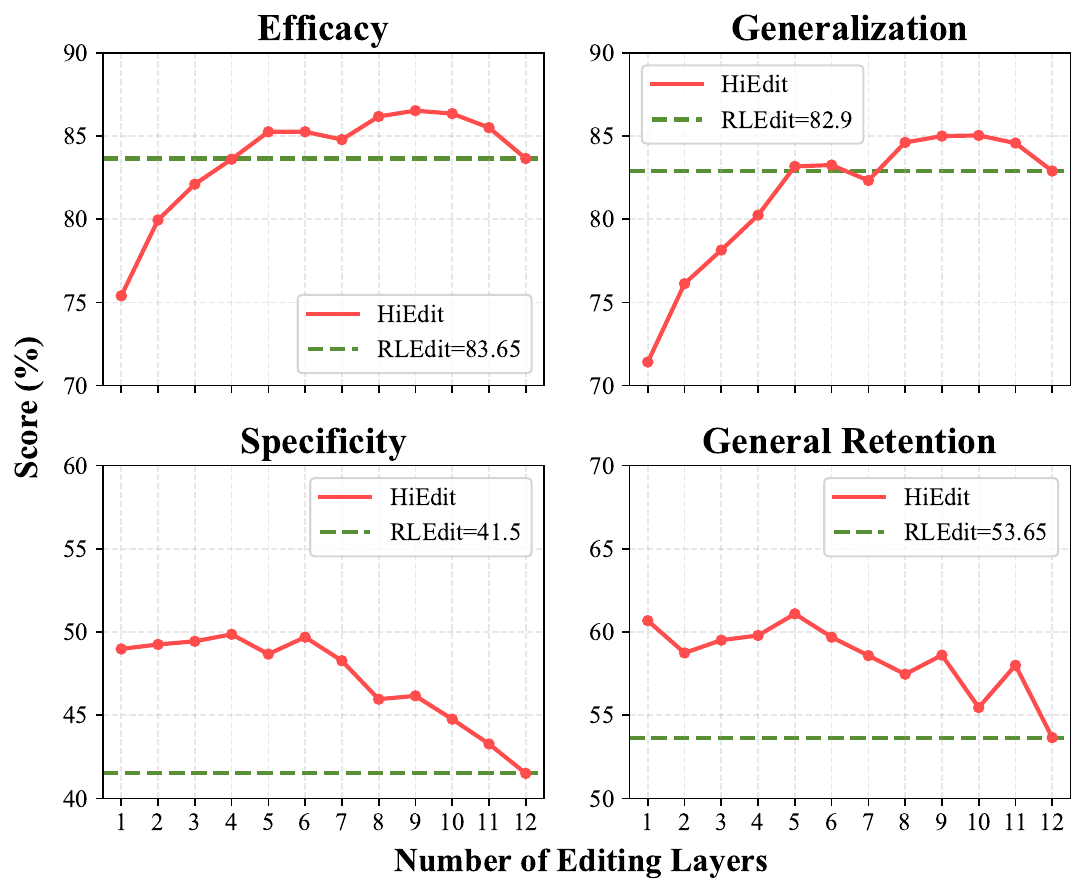}
  \caption {HiEdit's performance with varied number of editing layers.}
  \label{fig:n_layers}
\end{figure}

In Figure~\ref{fig:n_layers}, we presents a comparison of HiEdit's performance with varying numbers of editing layer $K$, highlighting the performance gap between HiEdit and RLEdit under different $K$ settings. The hyperparameter $K$ ranges from 1 to 12, with $K=12$ making HiEdit equivalent to RLEdit. 

The experimental results show that: with a smaller $K$, the Efficacy and Generalization of editing decrease, but Specificity and General Retention are preserved due to fewer perturbations in LLM parameters. Conversely, a larger $K$ enhances the Efficacy and Generalization but compromises Specificity and General Retention due to excessive parameter perturbation. Notably, when $K$ is less than or equal to half the number of influential layers, HiEdit achieves performance comparable to or even better than RLEdit, indicating that editing on dynamic and sparse LLM layers can effectively preserve existing knowledge and facilitate the integration of new knowledge.

\section{Computational Cost Analysis}
\label{app:co}

\paragraph{Parameter Counts.} We compare the parameter counts of different hypernetwork implementation styles for the HiEdit introduced high-level hypernetworks, the original low-level hypernetworks, and the Llama-3-8B model. The results are summarized in Table~\ref{tab:count}.

\begin{table}[h]
  \centering
  \scalebox{0.7}{
  \begin{tabular}{cccc}
  \toprule
  \textbf{Style} & \textbf{High-level} & \textbf{Low-level} & \textbf{Llama-3-8B} \\
  \midrule
  MEND & 4.76M & 152.83M & 8.03B \\
  MALMEN & 4.76M & 142.47M & 8.03B \\
  \bottomrule
  \end{tabular}
  }
  \caption{Comparison of parameter counts between the HiEdit introduced high-level hypernetworks, the original low-level hypernetworks, and the Llama-3-8B model.}
  \label{tab:count}
\end{table}

HiEdit introduces a high-level hypernetwork for adaptive layer selection prior to editing, adding architectural complexity. However, the additional high-level hypernetwork’s parameter count is minimal—only 3.12\% of the original low-level hypernetwork’s parameters. Furthermore, the combined parameter count of both high-level and low-level hypernetworks is merely 1.96\% of the LLM’s total parameters. This demonstrates that HiEdit effectively addresses memory consumption and scalability concerns while maintaining its editing capabilities.

\begin{figure}[h]
  \centering
  \includegraphics[width=\linewidth]{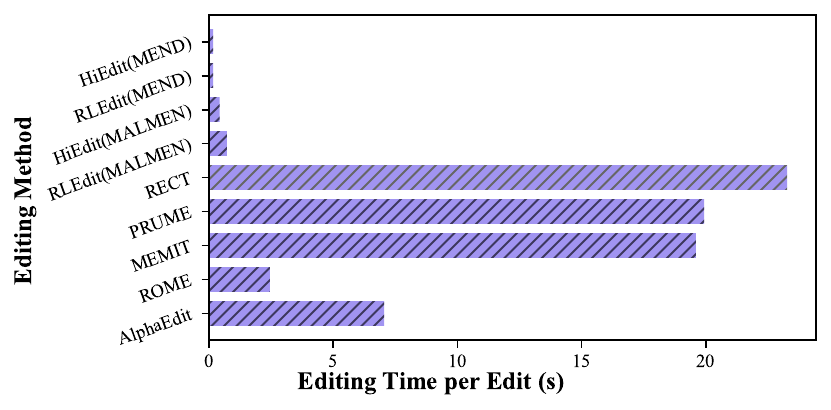}
  \caption {Comparison of editing time between HiEdit and various model editing methods.}
  \label{fig:time}
\end{figure}

\paragraph{Editing Efficiency.} We also compare the average editing time required for editing knowledge samples across different model editing methods. As shown in Figure~\ref{fig:time}, hypernetwork-based methods exhibit superior editing speed compared to other approaches, consistent with findings in prior research \citep{li2025reinforced}. HiEdit further enhances editing efficiency by reducing the number of editing layers through its adaptive layer selection mechanism. This reduction not only improves editing precision but also significantly shortens editing time compared to other hypernetwork-based methods.

\section{Case Study}
\label{app:case}

\begin{figure*}[ht]
  \centering
  \includegraphics[width=\linewidth]{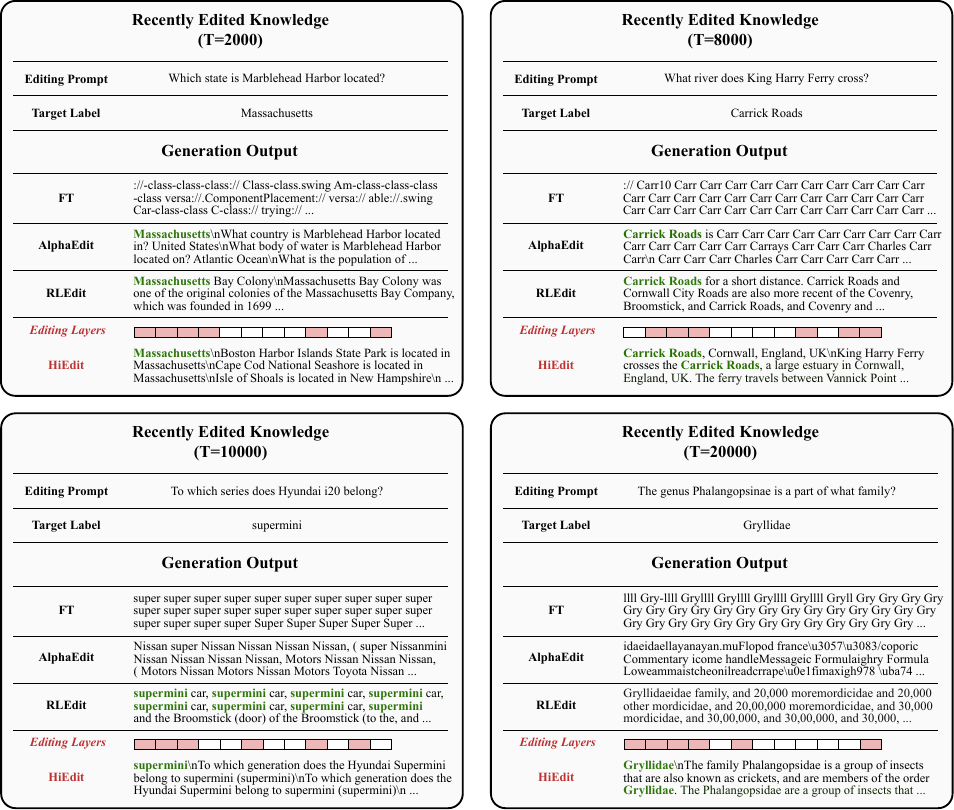}
  \caption {Samples of the recently edited knowledge instances.}
  \label{fig:case2}
\end{figure*}

\begin{figure}[h]
  \centering
  \includegraphics[width=\linewidth]{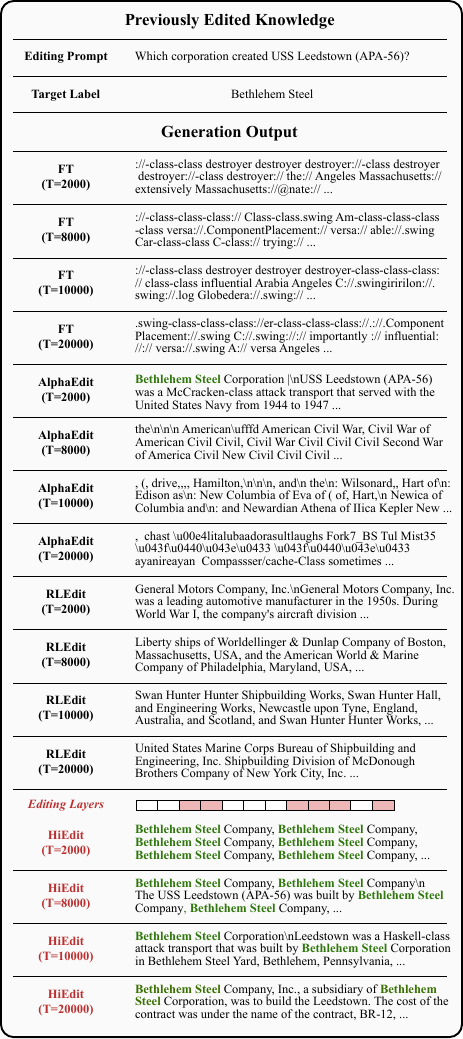}
  \caption {A sample of the initially edited knowledge instances.}
  \label{fig:case1}
\end{figure}

In this section, we present some cases to demonstrate how HiEdit and other competitive baseline methods (such as FT, AlphaEdit, and RLEdit) generate knowledge from earlier (Figure~\ref{fig:case1}) and recent (Figure~\ref{fig:case2}) edits after 2,000, 8,000, 10,000, and 20,000 sequential model edits on the ZsRE dataset and Llama-3-8B. We also examine which model layers HiEdit selects during editing, highlighting them in pink \textcolor{pink}{$\blacksquare$}.

The results indicate that after a large number of sequential edits, baseline methods often suffer from output collapse or edit failures, particularly affecting earlier edits. In contrast, HiEdit effectively selects differentiated model layers for editing distinct knowledge, maintaining high-quality output and successful updates for both earlier and recent edits, even after 20,000 edits. This underscores HiEdit's superior performance and robustness in lifelong model editing, as well as its scalability to accommodate ever-increasing editing scale.

\newpage

\end{document}